\documentclass[10pt,twocolumn,letterpaper]{article}

\usepackage[pagenumbers]{cvpr} 
\newcommand{\PAR}[1]{\vskip4pt \noindent{\bf #1~}}
\usepackage{algorithm}
\usepackage{amsmath}

\usepackage{tablefootnote}
\usepackage{threeparttable}
\usepackage{multirow}
\usepackage{array}
\usepackage{comment}
\newcolumntype{P}[1]{>{\centering\arraybackslash}p{#1}}
\usepackage{algpseudocode}
\usepackage[dvipsnames]{xcolor}

\definecolor{cvprblue}{rgb}{0.21,0.49,0.74}
\usepackage[pagebackref,breaklinks,colorlinks,citecolor=cvprblue]{hyperref}

\def\paperID{} 
\def\confName{CVPR}
\def\confYear{2024}

\title{Has Anything Changed? 3D Change Detection by 2D Segmentation Masks}

\author{Aikaterini Adam$^{1,2}$ \and Konstantinos Karantzalos$^2$ \and Lazaros Grammatikopoulos$^3$ \and Torsten Sattler$^1$\\ \\
$^1$ Czech Technical University in Prague\\
$^2$ National Technical University of Athens\\
$^3$ University of West Attica }

\begin{document}

\twocolumn[{%
\renewcommand\twocolumn[1][]{#1}%
\maketitle
\begin{center}
    \captionsetup{type=figure}
    \includegraphics[width=\linewidth,height=\textheight,keepaspectratio, trim={0.5cm 8cm 4cm 0cm},clip]{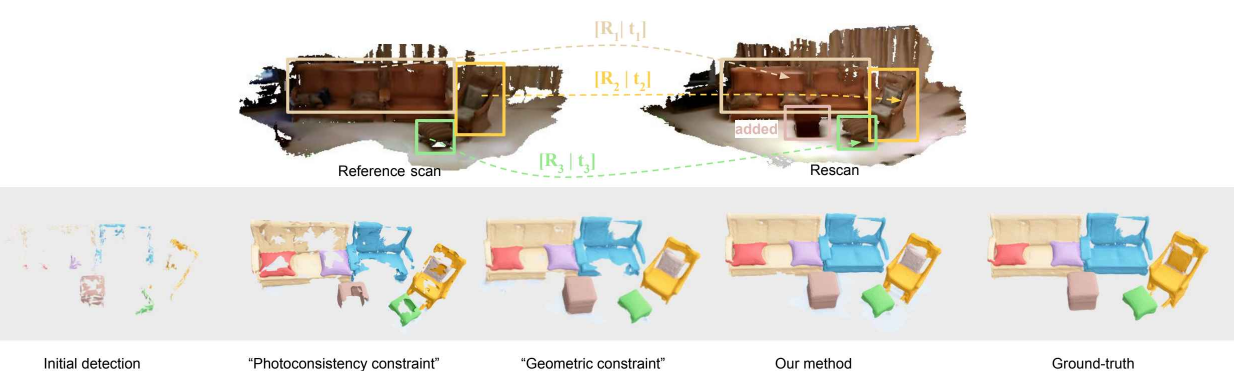} 
    \captionof{figure}{\textbf{Given two scans, a reference scan and a rescan we segment out any changed object (added, moved, or removed)}.  Firstly, we get an initial but incomplete set of detections, through render-and-compare. We then propagate the change to the whole object defined by a measure of constraint. Our method, using the “segmentation mask constraint” achieves SoTA performance, against the “photoconsistency constraint” and the “geometric constraint”.  The colors source from the instance segmentation of the scene, towards a more clear visualization.}\label{fig:teaser1}
\end{center}%
}]


\begin{abstract}
As capturing devices become common, 3D scans of interior spaces are acquired on a daily basis. Through scene comparison over time, information about objects in the scene and their changes is inferred. This information is important for robots and AR/VR devices, in order to operate in an immersive virtual experience. We thus propose an unsupervised object discovery method that identifies added, moved, or removed objects without any prior knowledge of what objects exist in the scene. We model this problem as a combination of a 3D change detection and a 2D segmentation task. Our algorithm leverages generic 2D segmentation masks to refine an initial but “incomplete” set of 3D change detections. The initial changes, acquired through render-and-compare likely correspond to movable objects. The incomplete detections are refined through graph optimization, distilling the information of the 2D segmentation masks in the 3D space. Experiments on the 3Rscan dataset prove that our method outperforms competitive baselines, with SoTA results. Our code will become available at \url{https://github.com/katadam/ObjectChangeDetection}.
    
\end{abstract}

\section{Introduction}
Building 3D models of scenes is becoming easier
as the required hardware and software mature and begin to become available to regular consumers. With the advancements in affordable depth-sensing devices operating in everyday life (\eg Microsoft  Kinect~\cite{glocker2013real} and HoloLens, Apple's iPhone) 3D scans of indoor environments become more available on a daily basis, making 3D models as common as images soon~\cite{halber2019rescan,reizenstein2021common,deitke2023objaverse,collins2022abo}.
Such a wide release of 3D models opens up interesting new possibilities, especially as re-scans, \ie 3D representations of already captured scenes, become more accessible. Comparing scans over time tells us about how scenes change, how humans interact with scenes, or what objects are in a scene. Such knowledge is crucial for agents interacting with the physical environment, such as robots, and augmented and virtual reality (AR/VR) devices, as it enables fully immersive virtual experiences, \ie merging the digital and physical worlds. The application we are interested in in this work is unsupervised object discovery, where we aim to identify changed objects without any predefined notion of what an object is. Indeed, checking for scene consistency and detecting changes on an object level can lead to unsupervised 3D object discovery, without the need for labeled data.

The object discovery problem is modeled as a change detection task. Given two scans of the same environment captured at different times, detected changes between the scans are likely to correspond to movable objects. The task is entangled with challenges due to the (1) noisy scans introduced by the low-cost-scanning devices, (2) lack of large-scale data to train an end-to-end procedure, (3) small-scale changes that are often missed, and (4) different illumination conditions. 

The main solutions so far operate in a render-and-compare fashion on registered images~\cite{palazzolo2018fast} or the point clouds~\cite{ambrucs2014meta}. Checking for inconsistencies in image pairs leads to the 3D changed regions in~\cite{palazzolo2018fast}, while approaches operating in 3D compare voxel occupancy. However, such methods often miss small-scale changes and might not discover the full object, \eg, when it partially overlaps with its original position.  
To propagate changes to the whole object region, subsequent methods enforce photoconsistency~\cite{taneja2011image} or geometric consistency~\cite{adam2022objects} constraints. More specifically, the “geometric constraint”~\cite{adam2022objects} exploits the coherent movement of objects and propagates changes to the whole region undergoing the same rigid transformation. However, these constraints remain sensitive to computing the geometric transformations induced by moving objects and variations in the illumination conditions of the scene.

To overcome these pitfalls, we propose a new constraint for change propagation to the whole object. Our work is inspired by the latest research in “foundation models”~\cite{bommasani2021opportunities}, enabling zero-shot generalization to conditions beyond those met during training. The transient limited access to 3D representations (compared to 2D data) affects the proficiency of such large 3D pre-trained models.  On the other hand, due to the abundance of 2D data, 2D vision models achieve remarkable results. Among them, we find 2D segmentation algorithms~\cite{kirillov2023segment}, generalizing well to new open-set environments. We thus leverage segmentation masks as an alternative modality that better identifies object segments. The 2D segmentation masks~\cite {kirillov2023segment} typically segment (or oversegment) general objects, without being limited to a pre-defined list of objects (unlike semantic and instance segmentation algorithms). Having such segmentations could help simplify the problem by allowing us to obtain better segments and better determine which parts of the 3D scene moved together. In this paper, we investigate to which extent the usage of such 2D segmentation algorithms impacts the detection of changed objects in 3D scenes.

We thus model our problem as a combination of a 3D change detection and a 2D segmentation step. The proposed method takes two 3D scans captured at different times and the associated camera poses, and segments any changed (added, moved, or removed) 3D object between the two data captures. The method is composed of two main steps: change detection and object segmentation. During change detection, an initial set of moving points is detected via depth render-and-compare. As stated above, in small-scale changes, such as the ones happening in household settings, such a depth comparison discovers only parts of the moving objects, \ie mainly their boundaries. Thus, we introduce a new constraint towards the refinement of the initial changes. To propagate the detection to the full object, even in parts where the depth does not change drastically, we leverage the connectivity of the object, expressed by 2D segmentation masks. Hence, the initial detections serve as “seeds”. Then, the off-the-shelf segmentation mask introduces a 2D prior to the discovered objects and when distilled in the 3D space,  propagates the change to the whole object.

\PAR{Contributions.} We summarize our contribution as follows:

i) We introduce a novel unsupervised method for discovering and segmenting out changed objects in 3D scenes without any prior knowledge of the objects in the scene. The framework is fully unsupervised and does not encode any strong assumptions of what an object “looks like”.

ii) We build upon prior baselines, by introducing a new “segmentation mask constraint” for change propagation, based on the latest research in the large, pre-trained models. We analyze to which extent the “generic” 2D segmentation masks lead to correctly discovering the whole changed object. Our method is more robust when compared against the SoTA algorithm “geometric constraint”~\cite{adam2022objects}, since it is much simpler and does not rely on computing the rigid transformations in the scene, towards change propagation.

iii)  Our method significantly improves the performance on the 3RScan dataset~\cite{wald2019rio} when compared against competitive baselines~\cite{adam2022objects, taneja2011image}.  Our code will become publicly available at \url{https://github.com/katadam/ObjectChangeDetection}.

\begin{figure*}[ht!]
\centering
\includegraphics[width=\linewidth,height=\textheight,keepaspectratio, trim={0cm 6cm 3cm 0cm},clip]{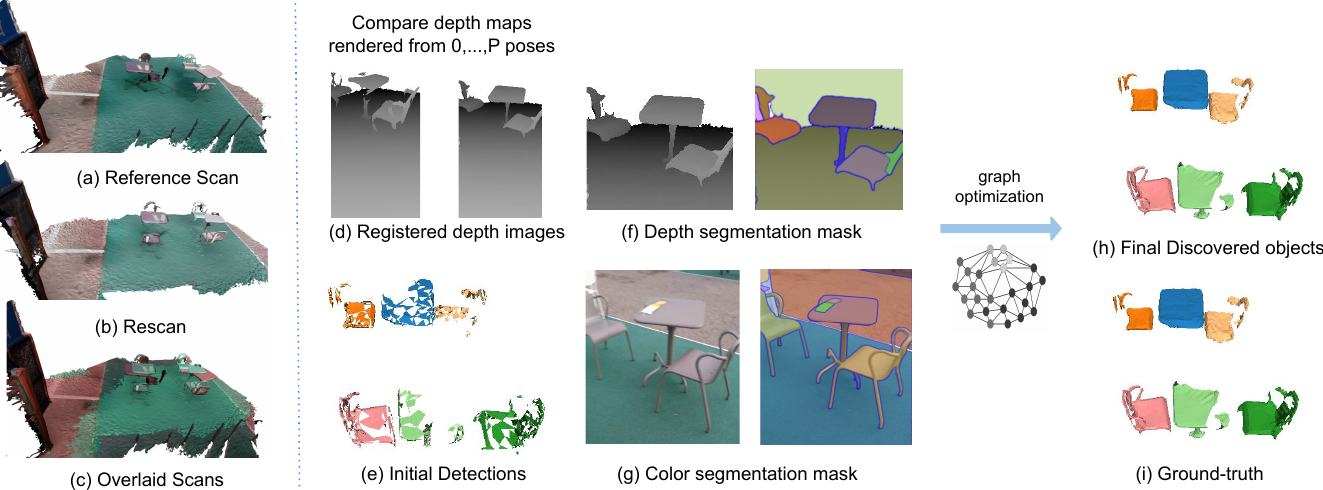} 
\caption{\textbf{Overview of our method}. Given two scans, (a) a reference scan, and (b) a rescan and the associate camera poses, we acquire an initial but incomplete set of detections via render-and-compare (d), (e). The initial detections are then refined, through our novel “segmentation mask constraint”, \ie a graph optimization~\cite{landrieu2017cut} propagating the change to the whole object sharing the same 2D mask (f),(g). The final results in (h) prove the effectiveness of the “segmentation mask constraint” when compared against the ground truth (i).}
\label{fig:pipeline}
\centering
\end{figure*}

\section{Related Work}\label{sec:related}

\PAR{Object Discovery via Motion Observation.} 

Discovering objects through motion cues is part of the ongoing research both in the 2D~\cite{bao2022discovering, bao2023object} and the 3D~\cite{ambrucs2014meta, ambrus2016unsupervised} domains. Ambrus \etal ~\cite{ambrucs2014meta} propose a change detection system through direct scene comparison. Occlusion checks are then performed to distinguish between real changes and segments occluded by dynamic objects. Contrary to our work, a full observation overlap is demanded, while our method handles partial observations when revisiting the scene. In a similar vein to our work, in~\cite{adam2022objects} changed objects are discovered through change propagation from the initial, incomplete set of detections to geometrically consistent regions. More specifically, the work is built upon the general assumption that objects move in a rigid way. Thus, it establishes correspondences between the two 3D scenes and then calculates the rigid transformations induced by the moving objects. Finally, change is propagated from the initial detections to all the parts undergoing the same rigid transformation, as rigid objects move coherently. Our work, on the contrary, enforces the constraints induced by the 2D segmentation masks towards change propagation, \ie propagates the change to all the parts sharing the same segmentation mask. Such segmentation masks typically segment general objects and thus introduce the concept of objects without having a pre-defined notion of what an object is.

\PAR{Change Detection.}
We model our problem as a 3D change detection and 2D segmentation task. Hence, our work is closely related to change detection, \ie the task of identifying differences between scenes or images. 

Similar to our initial detection step, ~\cite{taneja2011image, taneja2013city,palazzolo2017change, palazzolo2018fast, ulusoy2014image} target
change detection in urban scenes and environments through image comparison. Specifically, they discover changes in paired images by detecting depth and color inconsistencies.  As in our work, Taneja \etal~\cite{taneja2011image} further impose a photoconsistency constraint to propagate the change to partially missed areas. On the other hand, our work exploits the latest research in 2D large, pre-trained models towards change propagation. Finally, as our work targets change detection in open-set 3D environments, open-set 2D change detection between two “in the wild” images is presented in~\cite{sachdeva2023change}.

\PAR {SLAM in Dynamic Scenes.} Discovering dynamic objects in SLAM frameworks has been a long-standing issue in computer vision and robotics. Contrary to our unsupervised scheme, in~\cite{runz2018maskfusion} semantic scene understanding is exploited to map and track multiple objects. Fehr \etal~\cite {fehr2017tsdf} discover and reconstruct dynamic objects on a TSDF representation, while we work on supervoxels. Herbst \etal~\cite{herbst2014toward} explore dynamic object segmentation and modeling on desk scenes. Similar to our work, in ~\cite{finman2013toward} dynamic objects are discovered through object-level change detection. All these methods call for the direct observation (and tracking) of the motion of the object in front of the camera. Contrary to that, we are just given two observations (3D scans) that contain changes we need to discover. Thus, we explore a different problem.
\PAR{Segmentation in open-set environments.}  Segment Anything (SAM)~\cite{kirillov2023segment} is a large, 2D pre-trained model towards image segmentation. To enable zero-shot generalization to unseen image distributions and tasks, the authors designed a promptable segmentation task. Points, bounding boxes, masks, or even text serve as prompts. The model uses an MAE~\cite{he2017mask} pre-trained ViT-H~\cite{dosovitskiy2020image} image encoder.

{\PAR{Volumetric graph-cuts.}} Graph-cuts optimization in the 3D is exploited as part of different tasks. Similarly to our work, Taneja \etal~\cite{taneja2011image} deploy a graph-cut optimization for refining change detection results, on a voxelised representation. As in~\cite{adam2022objects}, we use a graph-cuts optimization on a supervoxel representation~\cite {papon2013voxel} towards the optimization of the initial changing regions. Landrieu \etal~\cite{landrieu2017structured} use a graph optimization on the point cloud to obtain spatially smooth semantic labels of 3D points.

\section{3D Object Discovery } \label{sec:method}
We introduce a new method that discovers changed objects between scenes recorded at long time intervals, without having a list of predetermined objects. Our method builds on top of “geometric constraint”~\cite{adam2022objects}, by introducing a novel “segmentation mask constraint”. As mentioned in \cref{sec:related}, “geometric constraint”~\cite{adam2022objects} uses the assumption that rigid objects move in the same rigid way, towards successful object discovery. Thus, it establishes 3D correspondences between the scan and the rescan and calculates the rigid transformations, induced by moving objects. The initial changes are then optimized by propagating the change to the whole region undergoing the same transformation.

As the “geometric constraint”~\cite{adam2022objects}, our conceptualization has three discrete steps
(1) identify regions of potential changes, referred to as “seeds”, (2) determine which parts of the scene likely belong to each other, \eg, the same object, (3) segment out objects by solving a graph cut optimization problem using the seed information for the unary and the pairwise information for the binary terms. The overview of our method is illustrated in \cref{fig:pipeline}. The main difference between our work and the “geometric constraint”~\cite{adam2022objects}, lies in the second step. The “geometric constraint”~\cite{adam2022objects} identifies an initial, but incomplete set of detections and propagates the changes to regions undergoing the same transformation. Contrary to that, our method refines the same initial set of detections, by introducing a novel constraint, the “segmentation mask constraint”, \ie by propagating the change to all the parts sharing the same segmentation mask. We thus leverage 2D segmentation masks from~\cite{kirillov2023segment} as an alternative modality that better identifies object segments, \ie parts of the scene that move together.

As a high-level overview of our method, we compute the “seeds” for the initial change detection by render-and-compare, unveiling regions with large depth differences. However, in indoor household environments, we mostly encounter small-scale changes. Thus, this initial set of detections is incomplete and has to be refined. To this end, we first discover the whole 2D objects of the seeds, through leveraging the segmentation masks from a  large 2D segmentation model, SAM~\cite{kirillov2023segment}. Since there is no direct supervision on the given dataset, our framework is perceived as an unsupervised method of discovering objects in 3D. The information from the 2D segmentation masks is finally aggregated in the 3D. As such, the 2D masks are back-projected in the 3D, and a graph optimization is applied to a supervoxel representation~\cite{papon2013voxel}. The graph optimization~\cite{landrieu2017cut} propagates the change from the seeds to all the regions sharing the same 2D masks, \ie to the whole object. A connected component analysis is then applied to form the finally discovered objects.

\subsection{Initial Set of Detections}\label{sec:seeds}

Following~\cite{adam2022objects},  given that the two scans (the reference scan $S$ and the rescan $R$) are aligned
\footnote{In line with the relevant literature \cite{halber2019rescan, wald2019rio, adam2022objects}, we do not perform any pre-processing of registering the scans with the rescans, as this is outside the scope of this research.}
we shall use the same set of poses to render the depth maps from both scans. The rendered maps are registered, and the image pairs can be directly subtracted. 
The subtracted pairs are thresholded with the help of~\cite{barron2020generalization}, as in~\cite{adam2022objects}. 
Backprojecting the thresholded regions in the 3D  leads to the initial set of detections, \ie regions undergoing a large depth difference and could unveil changed objects. More details are provided in~\cite{adam2022objects}.

\PAR{Supervoxels.} Similarly to~\cite{adam2022objects}, our method operates on the supervoxel representation of the 3D space~\cite{papon2013voxel}. We represent our 3D scenes with the supervoxels since this structure offers a coarse over-segmentation of the scene while preserving object boundaries. Supervoxels are a much more meaningful representation for scene segmentation/object discovery since they consider spatial, color, and geometric characteristics versus the plain spatial information constraint of the voxel representation. In this step, we compute the supervoxel representation of our 3D scan, along with the “changing supervoxels”, \ie supervoxels including changing points. These initial detections of changing supervoxels, serve as the “seeds” of change since the initial set of detections is incomplete. The changes are then propagated to the whole object, through a graph optimization step, extending the change to all the parts of the scene, sharing the same segmentation mask.

\subsection{2D Segmentation Masks}\label{sec:segme_anything}
From the first step, we have a set of initial detections (changing supervoxels) belonging to potentially changing objects/parts of the scene. As outlined above, not all the parts of each object that underwent a change in pose will be detected in the first stage. In the second stage, we thus aim to determine which parts of the 3D model/scene belong together.

Prior works~\cite{adam2022objects, taneja2011image} use cues such as color and transformation consistency to determine which parts of the 3D model belong to the same object. More specifically, the “geometric constraint”~\cite{adam2022objects} deploys a geometric consistency measure to discover parts of the scene that moved together. On the other hand, we want to revisit the use of 2D segmentation masks for the tasks of segmenting out changed objects between two scans. We thus enforce a “segmentation mask constraint”. To extract the segmentation masks, we are considering the case where we do not know which objects are present in the scene. As such, we cannot use semantic/instance/panoptic segmentation approaches as they require a fixed set of classes. We thus turn to open-set generic segmentation, from large, pre-trained models~\cite{kirillov2023segment}.  The 2D segmentation masks from those models already introduce the concept of objects without having a pre-defined notion of what an object is. They are then distilled to the 3D to segment out the whole object. To this end, a graph optimization will enforce change propagation from the seeds, to the parts of the scene sharing the same SAM mask.

Towards that deployment, we calculate SAM segmentation masks. We then associate primitives (supervoxels) in 3D with each other depending on whether they belong to the same segment in one of the images. An alternative to that procedure would be to first compute a consistent labeling in 3D, \eg using~\cite{bhalgat2023contrastive}. We leave the interesting question of whether enforcing label consistency in 3D improves over using 2D labels without enforcing consistency for future work. For the primitives association in 3D, we project supervoxels into the images to identify which of them share the same SAM mask, \ie which of them belong together. This imposes the new “segmentation mask constraint” that enforces that supervoxels sharing the SAM mask (even in one image) should be labeled consistently. Supervoxels sharing the SAM masks is a very good indication that they are part of the same object. Thus, the graph optimization will attempt to propagate change from the initial changing supervoxels, to all the supervoxels sharing the same segmentation mask, since these regions are highly likely to belong to the same object and have been missed during the step of initial detection. 

To calculate the masks, we apply SAM~\cite{kirillov2023segment} out-of-the-box without any fine-tuning/training. The automatic mask generator is applied on both the raw color captured images of the RGB-D sequence as well as to the depth images. We deploy SAM on the depth images, with interesting results, even though it was never explicitly trained on depth maps. Indeed, as objects are defined by a relatively coherent depth compared to their background, we find the twist of applying SAM to depth images powerful enough to fulfill our requirements of discovering consistent 2D objects (as described by the 2D inferred segmentation masks).

With the help of SAM, we have at our disposal the 2D segmentation of masks on depth and color images. Given that the images are posed, along with the depth image, we can associate pixels with the corresponding 3D points, through back-projection. 
Thus, we can assign a SAM mask to each 3D point $p$ of the scene. Consequently, we can also assign a SAM mask for each supervoxel $V$, as the most frequent mask of the points $p \in V$. Given that we have calculated a SAM mask for each supervoxel, we can now associate these primitives (supervoxels), sharing the same SAM mask in each image.

\subsection{Optimization of the initial changes}

From steps (1) and (2), we compute the initial change detections and gather information about which parts of the scene are likely to belong together (\ie share the same segmentation mask). In the final part, we thus put together these pieces of information to segment out objects in 3D. Following a common approach in the literature~\cite{landrieu2017cut}, we cast this problem as a graph cut problem. As in “geometric constraint”~\cite{adam2022objects}, we deploy the optimization on a supervoxel representation. In this last step, the graph optimization distills the information from the SAM segmentation masks into the 3D space. Thus, it enforces that the change is propagated from the “seeds” to the whole parts of the scene sharing the same SAM masks, and as a consequence to the whole object. The optimization is performed on the rescan since the task aims to detect what objects have changed when revisiting the same place, \ie what has changed in the rescan.

Keeping the same configuration as in “geometric constraint”, we deploy the optimization to the undirected graph $\mathcal{G}=({V},{E},{w})$. Each node  $v_{i}\in{V}$ corresponds to a supervoxel and the edges $e_{i}\in{E}$ with $w$ weight connect adjacent supervoxels. The goal of the optimization is to calculate an optimized labeling $P^{*}$, assigning a changing or non-changing label to each supervoxel . The optimized labeling $P^{*}$ should remain as close as possible to the initial solution $P$ (in terms of the supervoxels initially labeled as changing), while enforcing the condition that parts of the same object are labeled the same. Thus, $P^{*}$ should stay close to the initial change detection while enforcing that supervoxels sharing the same segmentation mask, are assigned the same label (whether changing or non-changing). 

The graph optimization is solved using the “Generalized Minimal Partition” (GMP) problem~\cite{landrieu2017cut}, which can be perceived as a continuous-space version of the Potts energy model: 

\vspace{-10pt}
\begin{equation} 
P^{*} \in {\underset{Q\in\Omega}{\arg\min}} \{\Phi(P,Q)+\lambda\Psi(Q)\} \enspace .
\label{eq:optimized labeling}\enspace 
\end{equation} 
In~\cref{eq:optimized labeling}, the first term $\Phi$ stands for the fidelity term, and the second term $\Psi(Q)$, for the penalty function. $\Omega$ stands for the search space of the final solution.

The initial labeling $P$ is calculated as in “geometric constraint” and the same hyperparameters (\eg Kullback-Leibler fidelity function~\cite{kullback1951information}) are used, to fairly assess the impact of our newly introduced “segmentation mask constraint”. From the above, it is clear that the basic difference between our novel framework and “geometric constraint” is the penalty term.  
The penalty term is trying to enforce the new piece of information, \ie in our case, the segmentation mask induced by SAM, with $\lambda$ being the regularized parameter between the two terms. The penalty term can be decomposed in \cref{eq:penalty}: 
\vspace{-2pt}
\begin{equation} 
\label{eq:penalty}
\Psi(Q) = \sum_{(i, j) \in E{w_{i, j}}} \phi(v_i - v_j) 
\enspace ,
\end{equation} where $w_{i, j}$ stands for the weight across the edge $E_{i, j}$. Here, we used a constant weight $w_{i, j}$ across all edges. The term $\phi(v_i - v_j)$ is a piece-wise constant function, based on the Potts model, as described in \cref{eq:Potts}:
\vspace{-2pt}
\begin{equation} \label{eq:Potts}
\phi(v_i - v_j)  = \begin{cases}
  1 & \text{ if  ${v_{i},v_{j}}$ same segment. mask}\\
  0 & \text{otherwise} 
  \end{cases} \enspace .
  \vspace{-2pt}
\end{equation}

\cref{eq:Potts} assigns a high cost to the edges connecting supervoxels ${v_{i},v_{j}}$ sharing the same segmentation masks (as induced by SAM). Hence, parts of the scene, sharing the same segmentation mask, are penalized if they do not get the same final labeling (changing or non-changing).

\cref{eq:optimized labeling} is non-convex due to the Potts penalty term. Thus, its global minimum cannot be realistically solved for large point clouds. We thus use the $\ell_0$-cut pursuit algorithm introduced in~\cite{landrieu2017cut} that quickly approximates a solution with a few graph-cut iterations.

\begin{figure*}[htb!]
\centering
\includegraphics[width=\linewidth,height=\textheight,keepaspectratio, trim={0cm 8.5cm 8.7cm 0cm},clip]{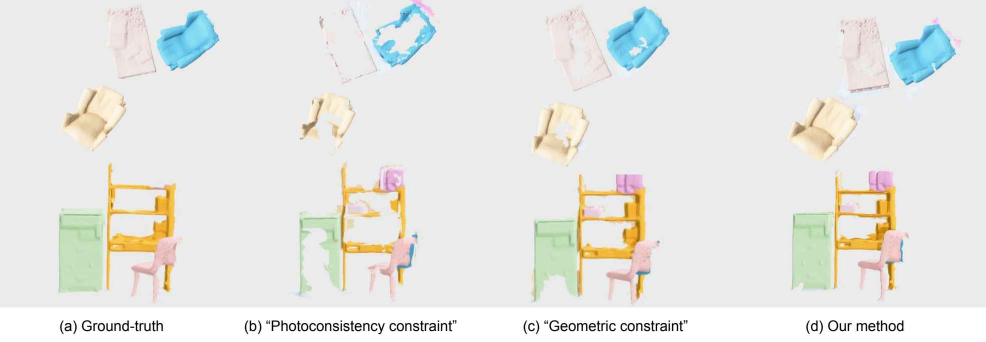} 
\caption{\textbf{Comperative results between the different change propagation constraints.} (a) Ground-truth, (b) “Photoconsistency constraint”~\cite{taneja2011image}, (c) “Geometric constraint”~\cite{adam2022objects}, and (d) Our novel “segmentation mask constraint” results. The objects discovered from the “segmentation mask constraint” are much more complete. The colors source from the instance segmentation of the scene, for a more clear visualization. }
\label{fig:comparison}
\centering
\end{figure*}

\begin{table*}[t!]
    \centering
    \begin{tabular}{P{60mm}P{17mm}P{17mm}P{17mm}P{14mm}P{4mm}}
    \toprule
    \textbf{Method}                                    & \textbf{Recall@0.20} & \textbf{Recall@0.25}  & \textbf{Recall@0.50} & \textbf{AP@0.25} \\
    \midrule
    Palazzolo \etal~\cite{palazzolo2018fast}            &     54.23 & 40.00 & 31.10
    &  9.86\\
    “Photoconsistent” Taneja \etal~\cite{taneja2011image}                        &    48.10 & 33.07
    & 28.78 &  9.67\\
    “Geom. constraint” Adam \etal~\cite{adam2022objects}                     &    68.40 &  61.83 & 42.54  &  12.16  \\ 
    \midrule
    Ours [Geom. + SAM depth] &71.34  &  64.80 &  \textbf{44.24} & \textbf{14.04} \\
    Ours [Geom. + SAM color] & \textbf{74.52} & \textbf{65.70} &  43.30 & 12.94\\ 
    Ours [Geom. + SAM depth + SAM color] & \textbf{74.52} & \textbf{65.70}  & 43.30 & 12.94\\
    \midrule
    Ours [SAM depth]  &71.34  &  64.80 &  \textbf{44.24} & \textbf{14.04}\\ 
    Ours [SAM color] & \textbf{74.52} & \textbf{65.70} & 43.30 & 12.94\\
    Ours [SAM depth + SAM Color]  & \textbf{74.52} & \textbf{65.70}  & 43.30 & 12.94\\
    \bottomrule
    \end{tabular}
    \caption{\textbf{Recall@kIoU and AP@25 for our discovered 3D objects vs baselines on 3Rscan val. set.} We report the recall at IoU thresholds 0.25 and 0.5 as widely used~\cite{qi2019deep}. Similar to~\cite{adam2022objects}, we also report metrics at 0.20. For the relative comparison between the methods, we report AP@0.25. Our novel “segmentation mask constraint” outperforms a set of competitive baselines.}
    \label{Tab:results baselines}
\end{table*}

\section{Experimental Evaluation}\label{experimental}
\PAR{Datasets.}
Following ~\cite{adam2022objects}, we evaluate the efficiency of our proposed method on the 3Rscan dataset. A thorough discussion on datasets, highlighting the lack of suitable datasets for our task, can be found in the supplementary material (SM). 

It is important to note here that the 3Rscan dataset~\cite{wald2019rio} is designed for benchmarking instance relocalization. Previous work~\cite{adam2022objects} though sets it up towards 3D change detection/object discovery by exploiting the changes naturally recorded in the data. However, the annotation for this task is not fully complete. As shown in~\cite{adam2022objects}, there is no exhaustive recording of all the changed objects. 

The 3Rscan dataset holds several maps of indoor scenes collected at various times and where objects change in between recordings. Each scene comprises a reference scan of an individual room and multiple partial rescans of the room, recorded after changes. Along with the scans stored as 3D meshes, the dataset provides the RGB-D image sequences, the associated corresponding camera poses, instance segmentation for each mesh, and change annotation. Change annotation refers to changed objects (expressed as faces in the 3D mesh). For a fair comparison, we use the same splits defined in~\cite{adam2022objects} and we evaluate our method on the validation set. The validation split includes 47 different scenes (and thus reference scans), 110 rescans, and covers a wide array of mostly household scenes such as offices, living rooms, kitchens, and bedrooms. 

\PAR{Metrics.}We rigorously evaluate our method using metrics capturing the success of the object discovery/detection (of 3D changed objects). As in~\cite{adam2022objects}, we use recall@kIoU, as our main metric. The recall@kIoU = $\frac{1}{N}\sum_{i=0}^{N}u_{i}$, where $u_{i} \in \{0,1\}$ is set to $1$ if the IoU score for the $i-th$ box is greater than $k$, otherwise $0$. Thus, the metric returns the percentage of correctly retrieved changed objects at kIoU, out of all the changed objects $N$. 

We use the recall@kIoU since it is a metric robust to False Positives. As stated above, the annotation of changed objects in the dataset is not exhaustive, thus the metrics considering Precision/ False Positives are suspect to the incomplete annotation. On the other hand, recall@kIoU, as in~\cite{adam2022objects}, is an appropriate metric to evaluate the percentage of correctly retrieved objects, while penalizing the overdetections with IoU. 

Concerning the thresholds k, we use the IoU@0.25 and IoU@0.50, which are largely used in 3D object detection~\cite{brazil2023omni3d}, as established in~\cite{qi2019deep}. For the sake of completeness, we also tabulate results for the IoU@0.20, as in~\cite{adam2022objects}. Finally, for the relative comparison between the baselines and not as an absolute metric due to its sensitivity in False Positives, we also evaluate the different baselines using AP@0.25.

\PAR{Methods in comparison.} Our paper introduces a novel “segmentation mask constraint”, propagating change from the seeds to all the regions sharing the same SAM mask. We want to measure the impact of our newly introduced “segmentation mask constraint”
versus the constraints imposed by the geometric transformations~\cite{adam2022objects} and  photoconsistency~\cite{taneja2011image}. Thus, the main baselines for comparison are the “geometric constraint”~\cite{adam2022objects} and the “photoconsistency constraint”~\cite{taneja2011image}. We also include~\cite{palazzolo2018fast}, which is equivalent to our initial detection step.  Similarly to~\cite{adam2022objects} we discard from the comparison dynamic SLAM methods~\cite{runz2018maskfusion,finman2013toward,herbst2011toward,fehr2017tsdf}, since our method is complementary to SLAM-based techniques, as explained in \cref{sec:related}. 
We also do not include~\cite{ambrucs2014meta}, as the method demands a full observation overlap (between the scan and the rescan) and does not handle the partial observations of the 3Rscan dataset. 
Methods presented in~\cite{langer2020robust,mason2012object} integrate semantics and thus fall outside the scope of this work, which is trying to discover changed objects without any predefined notion of objects. Finally, we do not include~\cite{langer2017fly}, as it focuses on discovering only the added objects in the scene, while our approach segments out all the changed objects (added, moved, or removed).


We present our results when the “segmentation mask constraint” utilizes the 2D segmentation masks inferred on depth images (SAM depth) and on the color images (SAM color), and when this information is combined (SAM depth + SAM color). Moreover, we examine how adding the “geometric constraint” on top of the “segmentation mask constraint” affects the performance of the method (Geometry + SAM depth, Geometry + SAM color, Geometry + SAM depth + SAM color).

Finally, we also present the results of our method, with a different algorithm introducing off-the-shelf masks, \ie Mask-RCNN~\cite{he2017mask}. We apply the 2D mask-RCNN object detector, trained on the COCO dataset~\cite{lin2014microsoft}, on the color images of each rescan, without any extra fine-tuning. We thus replace our “segmentation mask constraint” from SAM, with the instance segmentation masks of mask-RCNN. The segmentations are used, similarly to SAM masks. A mask-RCNN label  $S_{i}$ is assigned to each supervoxel $v_i$, as the most common label of points that fall within the supervoxel. The binary term of the graph is then computed as stated in Equation \ref{eq:7}:

\begin{equation} \label{eq:7}
\vspace{-2pt}
\phi(v_i - v_j)  = \begin{cases}
  1 & \text { if ${v_{i},v_{j}}$ same mask-RCNN}\\
  0 & \text{ otherwise}
  \end{cases}
\enspace .
\vspace{-2pt}
\end{equation}

\PAR{Experimental Results.}
\cref{Tab:results baselines} presents the results for our novel method versus the proposed baselines. By observing \cref{Tab:results baselines}, it is clear that our method outperforms the most competitive baseline “geometric constraint” in all metrics. Introducing the constraints from the 2D segmentation masks on color images (SAM color), our method performs +6.12\% better compared to the SoTA for the recall@0.20, +3.87\% for the recall@0.25 and +0.76\% better in terms of the recall@0.50. Similarly, when SAM is deployed on depth images (SAM depth), the recall@0.20 is bootstrapped by +2.94\%, the recall@0.25 by +2.97\%, and the recall@0.50 by +1.7\%, achieving state-of-the-art performance. Considering AP@0.25, the absolute metrics are rather low, due to the incomplete ground-truth annotation mentioned above. However, the metrics showcase the improved performance of our novel “segmentation mask constraint” versus a bunch of competitive baselines.

\begin{table*}[ht!]
    \centering
    \begin{tabular}{P{40mm}P{17mm}P{17mm}P{17mm}P{14mm}P{4mm}} 
    \toprule
    \textbf{Method}                                    & \textbf{Recall@0.20} & \textbf{Recall@0.25}  & \textbf{Recall@0.50} & \textbf{AP@0.25} \\
    \midrule
    Mask-RCNN masks~\cite{he2017mask}                     &  52.96  & 44.96  & 35.20  &   6.00 \\ 
    \midrule
    Ours [SAM depth] &71.34  &  64.80 &  \textbf{44.24} &  \textbf{14.04}\\
    Ours [SAM color] & \textbf{74.52} & \textbf{65.70} & 43.30 & 12.94\\
    \bottomrule
    \end{tabular}   
    \caption{\textbf{Recall@kIoU and AP@25 for our discovered 3D objects vs “mask-RCNN masks” on the 3Rscan val. set.} We report the recall at IoU thresholds 0.25 and 0.5 as widely used~\cite{qi2019deep}. Similar to~\cite{adam2022objects}, we also report metrics at 0.20. For the relative comparison between the methods, we report AP@0.25. Our “segmentation mask constraint” outperforms the ablation baseline, using mask-RCNN masks.}
    \label{Tab:results ablation}
\end{table*}

\begin{figure}[htb!]
\centering
\includegraphics[keepaspectratio, scale = 1.3, trim={0cm 10.0cm 21.2cm 0cm},clip]{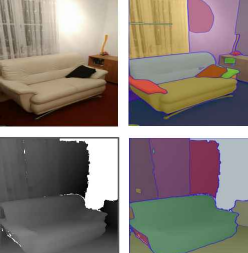} 
\caption{\textbf{The 2D segmentation masks~\cite{kirillov2023segment} when applied on color and depth images}. Objects tend to have consistent depth versus the background, thus leading to more uniform 2D segmentation masks on the depth images.} 
\label{fig:sam}
\centering
\vspace{-6mm}
\end{figure}

The results are validated visually in \cref{fig:comparison}, depicting that the “segmentation mask constraint” leads to “fuller” (\ie more complete) detections of the 3D objects. In other words, the “segmentation mask constraint” is more effective in propagating change from the seeds to the whole object sharing the same SAM mask, when compared to the “geometric constraint”. This could be potentially attributed to the noise in establishing correspondences in 3D and therefore, to poor results in computing rigid transformations and their inliers (needed for the “geometric constraint” as explained in the \cref{sec:related}). On the other hand, the “segmentation mask constraint” is conceptually a simpler method. Also, the segmentation masks induced by a large pre-trained model seem to be more robust to noise and when distilled in the 3D lead to superior results. Comparing the results when combining information from the segmentation masks on color and depth images, the performance seems to be capped by the information from the color 2D segmentation masks. As such, these sources of information (segmentation masks in depth and color) do not seem to be complementary. \cref{fig:sam} illustrates the inferred segmentation masks for the same part of the scene, in color and depth format. While statistically, the color images lead to a greater number of 2D segmentation masks, our method seems to be robust towards the “over-segmentation” from generic masks, with both methods (SAM color + SAM depth) performing almost the same.

\PAR{Ablation Study.}
From \cref{Tab:results baselines} it is clear that adding the “geometric constraint” on top of our newly introduced “segmentation mask constraint” does not help our method perform better. The performance of the Geometry+SAM depth is capped by the performance of the SAM depth baseline. Similarly, the performance of the Geometry+SAM color is capped by the performance of the SAM color baseline. Intuitively, this means that the “geometric constraint” propagates the change to regions already discovered by the “segmentation mask constraint” and thus, does not add any extra performance in the metrics. Also, this expresses that the “segmentation mask constraint” achieves
better results with a simpler method compared to the “geometric constraint” since our method does not require matching in 3D, and transformation estimation. When compared with another off-the-shelf method for mask segmentation in the images (mask-RCNN masks), it is clear that our method leveraging SAM~\cite{kirillov2023segment} performs significantly better. This is attributed to the lack of mask-RCNN's ability to generalize in unseen semantics during training. Hence, it proves the large advantage of distilling 2D large, pre-trained models into the 3D space. 

\PAR{Limitations.} Common failures of our method and~\cite{adam2022objects,taneja2011image} include wrong detections attributed to misalignments. Errors in the scan's registration can lead to significant depth differences in render-and-compare, and an erroneous set of initial changes. Our newly introduced “segmentation mask constraint” could be limited in case of benchmarks sourcing from hand-held devices. In these cases, there might be a deficiency of poses covering parts of objects lying next to the floor, which are often occluded (\eg the legs of the table). Limited coverage of such parts leads to fewer 2D masks and less successful change propagation. This limitation can be tackled by generating a variety of poses, ensuring that the whole scene is covered, to render new images. For more information please refer to the SM.

\section{Conclusion}
We present a new method, leveraging generic 2D segmentation masks towards 3D object discovery. Our presented method achieves state-of-the-art performance on the 3Rscan dataset, outperforming alternate approaches. The method deploys generic 2D segmentation masks, as a constraint to propagate change from an initial, incomplete set of detections, acquired by render-and-compare to the whole object. Our analysis proves the effectiveness of our newly introduced constraint towards a high-recall object discovery. Our method is unsupervised and does not demand prior information on what objects exist in the scene.

{
    \small
    \bibliographystyle{ieeenat_fullname}
    \bibliography{main}

\begin{thebibliography}{43}
\providecommand{\natexlab}[1]{#1}
\providecommand{\url}[1]{\texttt{#1}}
\expandafter\ifx\csname urlstyle\endcsname\relax
  \providecommand{\doi}[1]{doi: #1}\else
  \providecommand{\doi}{doi: \begingroup \urlstyle{rm}\Url}\fi

\bibitem[Adam et~al.(2022)Adam, Sattler, Karantzalos, and Pajdla]{adam2022objects}
Aikaterini Adam, Torsten Sattler, Konstantinos Karantzalos, and Tomas Pajdla.
\newblock Objects can move: 3d change detection by geometric transformation consistency.
\newblock In \emph{Eur. Conf. Comput. Vis.}, pages 108--124. Springer, 2022.

\bibitem[Ambru{\c{s}} et~al.(2014)Ambru{\c{s}}, Bore, Folkesson, and Jensfelt]{ambrucs2014meta}
Rare{\c{s}} Ambru{\c{s}}, Nils Bore, John Folkesson, and Patric Jensfelt.
\newblock Meta-rooms: Building and maintaining long term spatial models in a dynamic world.
\newblock In \emph{2014 IEEE/RSJ international conference on intelligent robots and systems}, pages 1854--1861. IEEE, 2014.

\bibitem[Ambrus et~al.(2016)Ambrus, Folkesson, and Jensfelt]{ambrus2016unsupervised}
Rares Ambrus, John Folkesson, and Patric Jensfelt.
\newblock Unsupervised object segmentation through change detection in a long term autonomy scenario.
\newblock In \emph{2016 IEEE-RAS 16th International Conference on Humanoid Robots (Humanoids)}, pages 1181--1187. IEEE, 2016.

\bibitem[Bao et~al.(2022)Bao, Tokmakov, Jabri, Wang, Gaidon, and Hebert]{bao2022discovering}
Zhipeng Bao, Pavel Tokmakov, Allan Jabri, Yu-Xiong Wang, Adrien Gaidon, and Martial Hebert.
\newblock Discovering objects that can move.
\newblock In \emph{IEEE Conf. Comput. Vis. Pattern Recog.}, pages 11789--11798, 2022.

\bibitem[Bao et~al.(2023)Bao, Tokmakov, Wang, Gaidon, and Hebert]{bao2023object}
Zhipeng Bao, Pavel Tokmakov, Yu-Xiong Wang, Adrien Gaidon, and Martial Hebert.
\newblock Object discovery from motion-guided tokens.
\newblock In \emph{IEEE Conf. Comput. Vis. Pattern Recog.}, pages 22972--22981, 2023.

\bibitem[Barron(2020)]{barron2020generalization}
Jonathan~T Barron.
\newblock A generalization of otsu’s method and minimum error thresholding.
\newblock In \emph{Eur. Conf. Comput. Vis.}, pages 455--470. Springer, 2020.

\bibitem[Bhalgat et~al.(2023)Bhalgat, Laina, Henriques, Zisserman, and Vedaldi]{bhalgat2023contrastive}
Yash Bhalgat, Iro Laina, Jo{\~a}o~F Henriques, Andrew Zisserman, and Andrea Vedaldi.
\newblock Contrastive lift: 3d object instance segmentation by slow-fast contrastive fusion.
\newblock \emph{arXiv preprint arXiv:2306.04633}, 2023.

\bibitem[Bommasani et~al.(2021)Bommasani, Hudson, Adeli, Altman, Arora, von Arx, Bernstein, Bohg, Bosselut, Brunskill, et~al.]{bommasani2021opportunities}
Rishi Bommasani, Drew~A Hudson, Ehsan Adeli, Russ Altman, Simran Arora, Sydney von Arx, Michael~S Bernstein, Jeannette Bohg, Antoine Bosselut, Emma Brunskill, et~al.
\newblock On the opportunities and risks of foundation models.
\newblock \emph{arXiv preprint arXiv:2108.07258}, 2021.

\bibitem[Boykov and Kolmogorov(2004)]{boykov2004experimental}
Yuri Boykov and Vladimir Kolmogorov.
\newblock An experimental comparison of min-cut/max-flow algorithms for energy minimization in vision.
\newblock \emph{IEEE Trans. Pattern Anal. Mach. Intell.}, 26\penalty0 (9):\penalty0 1124--1137, 2004.

\bibitem[Brazil et~al.(2023)Brazil, Kumar, Straub, Ravi, Johnson, and Gkioxari]{brazil2023omni3d}
Garrick Brazil, Abhinav Kumar, Julian Straub, Nikhila Ravi, Justin Johnson, and Georgia Gkioxari.
\newblock Omni3d: A large benchmark and model for 3d object detection in the wild.
\newblock In \emph{IEEE Conf. Comput. Vis. Pattern Recog.}, pages 13154--13164, 2023.

\bibitem[Collins et~al.(2022)Collins, Goel, Deng, Luthra, Xu, Gundogdu, Zhang, Vicente, Dideriksen, Arora, et~al.]{collins2022abo}
Jasmine Collins, Shubham Goel, Kenan Deng, Achleshwar Luthra, Leon Xu, Erhan Gundogdu, Xi Zhang, Tomas F~Yago Vicente, Thomas Dideriksen, Himanshu Arora, et~al.
\newblock Abo: Dataset and benchmarks for real-world 3d object understanding.
\newblock In \emph{IEEE Conf. Comput. Vis. Pattern Recog.}, pages 21126--21136, 2022.

\bibitem[Deitke et~al.(2023)Deitke, Liu, Wallingford, Ngo, Michel, Kusupati, Fan, Laforte, Voleti, Gadre, et~al.]{deitke2023objaverse}
Matt Deitke, Ruoshi Liu, Matthew Wallingford, Huong Ngo, Oscar Michel, Aditya Kusupati, Alan Fan, Christian Laforte, Vikram Voleti, Samir~Yitzhak Gadre, et~al.
\newblock Objaverse-xl: A universe of 10m+ 3d objects.
\newblock \emph{arXiv preprint arXiv:2307.05663}, 2023.

\bibitem[Dosovitskiy et~al.(2020)Dosovitskiy, Beyer, Kolesnikov, Weissenborn, Zhai, Unterthiner, Dehghani, Minderer, Heigold, Gelly, et~al.]{dosovitskiy2020image}
Alexey Dosovitskiy, Lucas Beyer, Alexander Kolesnikov, Dirk Weissenborn, Xiaohua Zhai, Thomas Unterthiner, Mostafa Dehghani, Matthias Minderer, Georg Heigold, Sylvain Gelly, et~al.
\newblock An image is worth 16x16 words: Transformers for image recognition at scale.
\newblock \emph{arXiv preprint arXiv:2010.11929}, 2020.

\bibitem[Fehr et~al.(2017)Fehr, Furrer, Dryanovski, Sturm, Gilitschenski, Siegwart, and Cadena]{fehr2017tsdf}
Marius Fehr, Fadri Furrer, Ivan Dryanovski, J{\"u}rgen Sturm, Igor Gilitschenski, Roland Siegwart, and Cesar Cadena.
\newblock Tsdf-based change detection for consistent long-term dense reconstruction and dynamic object discovery.
\newblock In \emph{2017 IEEE International Conference on Robotics and Automation (ICRA)}, pages 5237--5244. IEEE, 2017.

\bibitem[Finman et~al.(2013)Finman, Whelan, Kaess, and Leonard]{finman2013toward}
Ross Finman, Thomas Whelan, Michael Kaess, and John~J Leonard.
\newblock Toward lifelong object segmentation from change detection in dense rgb-d maps.
\newblock In \emph{2013 European Conference on Mobile Robots}, pages 178--185. IEEE, 2013.

\bibitem[Gadde et~al.(2017)Gadde, Jampani, Marlet, and Gehler]{gadde2017efficient}
Raghudeep Gadde, Varun Jampani, Renaud Marlet, and Peter~V Gehler.
\newblock Efficient 2d and 3d facade segmentation using auto-context.
\newblock \emph{IEEE Trans. Pattern Anal. Mach. Intell.}, 40\penalty0 (5):\penalty0 1273--1280, 2017.

\bibitem[Glocker et~al.(2013)Glocker, Izadi, Shotton, and Criminisi]{glocker2013real}
Ben Glocker, Shahram Izadi, Jamie Shotton, and Antonio Criminisi.
\newblock Real-time rgb-d camera relocalization.
\newblock In \emph{2013 IEEE International Symposium on Mixed and Augmented Reality (ISMAR)}, pages 173--179. IEEE, 2013.

\bibitem[Halber et~al.(2019)Halber, Shi, Xu, and Funkhouser]{halber2019rescan}
Maciej Halber, Yifei Shi, Kai Xu, and Thomas Funkhouser.
\newblock Rescan: Inductive instance segmentation for indoor rgbd scans.
\newblock In \emph{IEEE Conf. Comput. Vis. Pattern Recog.}, pages 2541--2550, 2019.

\bibitem[He et~al.(2017)He, Gkioxari, Doll{\'a}r, and Girshick]{he2017mask}
Kaiming He, Georgia Gkioxari, Piotr Doll{\'a}r, and Ross Girshick.
\newblock Mask r-cnn.
\newblock In \emph{Int. Conf. Comput. Vis.}, pages 2961--2969, 2017.

\bibitem[Herbst et~al.(2011)Herbst, Henry, Ren, and Fox]{herbst2011toward}
Evan Herbst, Peter Henry, Xiaofeng Ren, and Dieter Fox.
\newblock Toward object discovery and modeling via 3-d scene comparison.
\newblock In \emph{2011 IEEE International Conference on Robotics and Automation (ICRA)}, pages 2623--2629. IEEE, 2011.

\bibitem[Herbst et~al.(2014)Herbst, Henry, and Fox]{herbst2014toward}
Evan Herbst, Peter Henry, and Dieter Fox.
\newblock Toward online 3-d object segmentation and mapping.
\newblock In \emph{2014 IEEE International Conference on Robotics and Automation (ICRA)}, pages 3193--3200. IEEE, 2014.

\bibitem[Katsura et~al.(2019)Katsura, Matsumoto, Kawamura, Ishigami, Okada, and Kurazume]{katsura2019spatial}
Ukyo Katsura, Kohei Matsumoto, Akihiro Kawamura, Tomohide Ishigami, Tsukasa Okada, and Ryo Kurazume.
\newblock Spatial change detection using normal distributions transform.
\newblock \emph{ROBOMECH Journal}, 6\penalty0 (1):\penalty0 1--13, 2019.

\bibitem[Kirillov et~al.(2023)Kirillov, Mintun, Ravi, Mao, Rolland, Gustafson, Xiao, Whitehead, Berg, Lo, et~al.]{kirillov2023segment}
Alexander Kirillov, Eric Mintun, Nikhila Ravi, Hanzi Mao, Chloe Rolland, Laura Gustafson, Tete Xiao, Spencer Whitehead, Alexander~C Berg, Wan-Yen Lo, et~al.
\newblock Segment anything.
\newblock \emph{arXiv preprint arXiv:2304.02643}, 2023.

\bibitem[Kullback and Leibler(1951)]{kullback1951information}
Solomon Kullback and Richard~A Leibler.
\newblock On information and sufficiency.
\newblock \emph{The annals of mathematical statistics}, 22\penalty0 (1):\penalty0 79--86, 1951.

\bibitem[Landrieu and Obozinski(2017)]{landrieu2017cut}
Loic Landrieu and Guillaume Obozinski.
\newblock Cut pursuit: Fast algorithms to learn piecewise constant functions on general weighted graphs.
\newblock \emph{SIAM Journal on Imaging Sciences}, 10\penalty0 (4):\penalty0 1724--1766, 2017.

\bibitem[Landrieu et~al.(2017)Landrieu, Raguet, Vallet, Mallet, and Weinmann]{landrieu2017structured}
Loic Landrieu, Hugo Raguet, Bruno Vallet, Cl{\'e}ment Mallet, and Martin Weinmann.
\newblock A structured regularization framework for spatially smoothing semantic labelings of 3d point clouds.
\newblock \emph{ISPRS Journal of Photogrammetry and Remote Sensing}, 132:\penalty0 102--118, 2017.

\bibitem[Langer et~al.(2017)Langer, Ridder, Cashmore, Magazzeni, Zillich, and Vincze]{langer2017fly}
Edith Langer, Bram Ridder, Michael Cashmore, Daniele Magazzeni, Michael Zillich, and Markus Vincze.
\newblock On-the-fly detection of novel objects in indoor environments.
\newblock In \emph{2017 IEEE International Conference on Robotics and Biomimetics (ROBIO)}, pages 900--907. IEEE, 2017.

\bibitem[Langer et~al.(2020)Langer, Patten, and Vincze]{langer2020robust}
Edith Langer, Timothy Patten, and Markus Vincze.
\newblock Robust and efficient object change detection by combining global semantic information and local geometric verification.
\newblock In \emph{2020 IEEE/RSJ International Conference on Intelligent Robots and Systems (IROS)}, pages 8453--8460. IEEE, 2020.

\bibitem[Lin et~al.(2014)Lin, Maire, Belongie, Hays, Perona, Ramanan, Doll{\'a}r, and Zitnick]{lin2014microsoft}
Tsung-Yi Lin, Michael Maire, Serge Belongie, James Hays, Pietro Perona, Deva Ramanan, Piotr Doll{\'a}r, and C~Lawrence Zitnick.
\newblock Microsoft coco: Common objects in context.
\newblock In \emph{Eur. Conf. Comput. Vis.}, pages 740--755. Springer, 2014.

\bibitem[Mason and Marthi(2012)]{mason2012object}
Julian Mason and Bhaskara Marthi.
\newblock An object-based semantic world model for long-term change detection and semantic querying.
\newblock In \emph{2012 IEEE/RSJ International Conference on Intelligent Robots and Systems (IROS)}, pages 3851--3858. IEEE, 2012.

\bibitem[Palazzolo and Stachniss(2017)]{palazzolo2017change}
Emanuele Palazzolo and Cyrill Stachniss.
\newblock Change detection in 3d models based on camera images.
\newblock In \emph{9th Workshop on Planning, Perception and Navigation for Intelligent Vehicles at the IEEE/RSJ Int. Conf. on Intelligent Robots and Systems (IROS)}, 2017.

\bibitem[Palazzolo and Stachniss(2018)]{palazzolo2018fast}
Emanuele Palazzolo and Cyrill Stachniss.
\newblock Fast image-based geometric change detection given a 3d model.
\newblock In \emph{2018 IEEE International Conference on Robotics and Automation (ICRA)}, pages 6308--6315. IEEE, 2018.

\bibitem[Papon et~al.(2013)Papon, Abramov, Schoeler, and Worgotter]{papon2013voxel}
Jeremie Papon, Alexey Abramov, Markus Schoeler, and Florentin Worgotter.
\newblock Voxel cloud connectivity segmentation-supervoxels for point clouds.
\newblock In \emph{IEEE Conf. Comput. Vis. Pattern Recog.}, pages 2027--2034, 2013.

\bibitem[Park et~al.(2021)Park, Jang, Yoo, Lee, Kim, and Kim]{park2021changesim}
Jin-Man Park, Jae-Hyuk Jang, Sahng-Min Yoo, Sun-Kyung Lee, Ue-Hwan Kim, and Jong-Hwan Kim.
\newblock Changesim: Towards end-to-end online scene change detection in industrial indoor environments.
\newblock In \emph{2021 IEEE/RSJ International Conference on Intelligent Robots and Systems (IROS)}, pages 8578--8585. IEEE, 2021.

\bibitem[Qi et~al.(2019)Qi, Litany, He, and Guibas]{qi2019deep}
Charles~R Qi, Or Litany, Kaiming He, and Leonidas~J Guibas.
\newblock Deep hough voting for 3d object detection in point clouds.
\newblock In \emph{Int. Conf. Comput. Vis.}, 2019.

\bibitem[Reizenstein et~al.(2021)Reizenstein, Shapovalov, Henzler, Sbordone, Labatut, and Novotny]{reizenstein2021common}
Jeremy Reizenstein, Roman Shapovalov, Philipp Henzler, Luca Sbordone, Patrick Labatut, and David Novotny.
\newblock Common objects in 3d: Large-scale learning and evaluation of real-life 3d category reconstruction.
\newblock In \emph{Proceedings of the IEEE/CVF International Conference on Computer Vision}, pages 10901--10911, 2021.

\bibitem[Runz et~al.(2018)Runz, Buffier, and Agapito]{runz2018maskfusion}
Martin Runz, Maud Buffier, and Lourdes Agapito.
\newblock Maskfusion: Real-time recognition, tracking and reconstruction of multiple moving objects.
\newblock In \emph{2018 IEEE International Symposium on Mixed and Augmented Reality (ISMAR)}, pages 10--20. IEEE, 2018.

\bibitem[Sachdeva and Zisserman(2023)]{sachdeva2023change}
Ragav Sachdeva and Andrew Zisserman.
\newblock The change you want to see (now in 3d).
\newblock In \emph{Int. Conf. Comput. Vis.}, pages 2060--2069, 2023.

\bibitem[Sun et~al.(2023)Sun, Hao, Huang, Savarese, Schindler, Pollefeys, and Armeni]{sun2023nothing}
Tao Sun, Yan Hao, Shengyu Huang, Silvio Savarese, Konrad Schindler, Marc Pollefeys, and Iro Armeni.
\newblock Nothing stands still: A spatiotemporal benchmark on 3d point cloud registration under large geometric and temporal change.
\newblock \emph{arXiv preprint arXiv:2311.09346}, 2023.

\bibitem[Taneja et~al.(2011)Taneja, Ballan, and Pollefeys]{taneja2011image}
Aparna Taneja, Luca Ballan, and Marc Pollefeys.
\newblock Image based detection of geometric changes in urban environments.
\newblock In \emph{Int. Conf. Comput. Vis.}, pages 2336--2343. IEEE, 2011.

\bibitem[Taneja et~al.(2013)Taneja, Ballan, and Pollefeys]{taneja2013city}
Aparna Taneja, Luca Ballan, and Marc Pollefeys.
\newblock City-scale change detection in cadastral 3d models using images.
\newblock In \emph{IEEE Conf. Comput. Vis. Pattern Recog.}, pages 113--120, 2013.

\bibitem[Ulusoy and Mundy(2014)]{ulusoy2014image}
Ali~Osman Ulusoy and Joseph~L Mundy.
\newblock Image-based 4-d reconstruction using 3-d change detection.
\newblock In \emph{Eur. Conf. Comput. Vis.}, pages 31--45. Springer, 2014.

\bibitem[Wald et~al.(2019)Wald, Avetisyan, Navab, Tombari, and Nie{\ss}ner]{wald2019rio}
Johanna Wald, Armen Avetisyan, Nassir Navab, Federico Tombari, and Matthias Nie{\ss}ner.
\newblock Rio: 3d object instance re-localization in changing indoor environments.
\newblock In \emph{IEEE Conf. Comput. Vis. Pattern Recog.}, pages 7658--7667, 2019.

\end{thebibliography}
}

\clearpage
\maketitlesupplementary

\begin{table*}[htb!]
\begin{threeparttable}
\centering
\begin{tabular}{|p{40mm}p{10mm}p{15mm}p{15mm}p{15mm}p{15mm}p{10mm}p{20mm}|}
\hline
Method & Scans & Rescans & Instance Seg. & Annotation & Available & Real & environment \\ \hline  \hline
Langer et al.\cite{langer2020robust} &    5     &   31   & \textemdash  & \checkmark  & \checkmark  & \checkmark  & office  \\
Finman \etal~\cite{finman2013toward}             &   2   &   67  &  ?   &     ?   &  \textemdash      & \checkmark   & office \\ 
Langer \etal~\cite{langer2017fly}            &     1       &      4   &   \textemdash   &  \textemdash    &  \textemdash & \checkmark & office \\ 
Katsura \etal~\cite{katsura2019spatial}                         &    2     & 10+?     &  ? & ?  & \textemdash    & \checkmark  & corridor/hall \\ 
Herbst \etal~\cite{herbst2011toward} &     4    &    24  & \checkmark  &  \checkmark     & \textemdash\tnote{1}& \checkmark & tables  \\ 
Mason \etal~\cite{mason2012object} &  1       &  67    &  \textemdash  &   \textemdash    & \checkmark\tnote{2}& \checkmark & office \\ 
Ambrus \etal~\cite{ambrus2016unsupervised} &   1      &   88   & \textemdash  & \textemdash\tnote{3}  & \checkmark & \checkmark  & office\\ 
Fehr \etal~\cite{fehr2017tsdf} & 3 & 23 & \textemdash  & \textemdash   & \checkmark  & \checkmark & household\\ 
Wald \etal~\cite{wald2019rio} - 3Rscan &   478      &   1482   & \checkmark &  \checkmark  &\checkmark~\tnote{4}  & \checkmark & household\\ 
Halber \etal~\cite{halber2019rescan} - Rescan &   13      &    45  & \checkmark & \textemdash  &\checkmark   & \checkmark & household  \\
Park \etal~\cite{park2021changesim} - ChangeSim &  10       & 80     & \checkmark  & \checkmark  & \checkmark & \textemdash  & warehouse\\
Armeni \etal \cite{sun2023nothing} - NSS & 6        &   27   & \textemdash & \textemdash  & \checkmark & \checkmark & construction\\ \hline
\end{tabular}
\caption{ \textbf{Comparison of applicable datasets for indoor scene change detection/object discovery}.  [1] The provided URL is not valid, [2] Data available upon request, [3] Inconsistent and incomplete annotation, [4] With provided code from~\cite{adam2022objects}.}
{\label{Tab:datasets}}

\end{threeparttable}
\end{table*}

This is the supplementary material for our paper “Has Anything Changed? 3D Change Detection by 2D Segmentation Masks” . Continuing the dataset discussion in \cref{experimental}, we provide information on the available evaluation datasets. We also explain why 3Rscan~\cite{wald2019rio} is best tailored to our needs in \cref{sec:dat_suppl}. \cref{sec:implementation} presents the implementation details of our method. We present additional quantitative and qualitative results in \cref{sec:quantitative_suppl} and \cref{sec:qualitative_suppl}. We also provide a more thorough view of the ablation study and the edge cases of our method (as discussed in \cref{experimental}-Limitations).  

\section{Datasets}\label{sec:dat_suppl}
So far, an established dataset for evaluating 3D indoor change detection/object discovery does not exist. The majority of published works evaluate their methods on their own datasets, which might not be publicly available~\cite{finman2013toward,langer2017fly,katsura2019spatial,herbst2011toward} or not annotated appropriately~\cite{ambrus2016unsupervised, fehr2017tsdf, mason2012object}. \cref{Tab:datasets} summarizes relevant works either on 3D change detection or SLAM in dynamic scenes, along with the used datasets, their size, annotation, and availability information.


\begin{table*}[t!]
    \centering
    \begin{tabular}{P{40mm}P{18mm}P{18mm}P{17mm}P{17mm}}
    \toprule
    \textbf{Method} & \multicolumn{2}{c}{\textbf{Parameters}} & \textbf{Recall@0.25}  & \textbf{Recall@0.50}  \\
    & changing probability & non-changing  probability & & \\

    \midrule
     SAM depth &  1.0 &  0.0  &  64.80 &  44.24  \\
     SAM color &  1.0 &  0.0  &  65.70  &   43.30 \\
    SAM depth + SAM color &  1.0 &  0.0  &   65.70  & 43.30  \\

       \midrule
     \textbf{SAM depth} &  \textbf{0.8} &  \textbf{0.2}  &   \textbf{64.80} &  \textbf{44.24} \\
     \textbf{SAM color}&  \textbf{0.8} &  \textbf{0.2}  & \textbf{65.70}  & \textbf{43.30}  \\
    \textbf{SAM depth + SAM color} &  \textbf{0.8} &  \textbf{0.2}  &  \textbf{65.70} & \textbf{43.30}  \\
    
       \midrule
     SAM depth  &  0.7 &  0.3 &  64.80 &  44.24 \\
     SAM color  &  0.7 &  0.3 &  65.70  & 43.30  \\
    SAM depth + SAM color  &  0.7 &  0.3 &  65.70 & 43.30  \\

           \midrule
     SAM depth  &  0.6 &  0.4 &  64.86 & 41.60  \\
     SAM color  &  0.6 &  0.4 & 64.56 & 41.10  \\
    SAM depth + SAM color  &  0.6 &  0.4 &63.95 & 41.41 \\

       \midrule
     SAM depth  &  0.5 &  0.5 &  18.67 & 17.46 \\
     SAM color &  0.5 &  0.5 &18.52 & 17.30 \\
    SAM depth + SAM color &  0.5 &  0.5 &18.52 &  17.30
    \\
  
    \bottomrule
    \end{tabular}
    \caption{\textbf{Recall@kIoU for different hyperparameters settings.} We report the recall at IoU thresholds 0.25 and 0.5 as widely used~\cite{qi2019deep}. We measure the robustness of our method against different hyperparameters, explained in \cref{eq:soft labeling 1} and \cref{eq:soft labeling 2}. The hyperparameters in bold are the ones used for our method in the main paper.  Our method starts showing sensitivity only when the hyperparameters are tuned with edge values.}
    \label{Tab:results robust}
\end{table*}

Discussing \cref{Tab:datasets}, the authors of \cite{langer2020robust} create their own dataset for evaluating changed objects in the scene in an office environment. The dataset is rather limited (smaller and less diverse) compared to 3Rscan (5 scenes vs. 478, and 31 rescans vs. 478, respectively).
Also, it only focuses on small objects from the YCB dataset \cite{gadde2017efficient}.
The dataset of ~\cite{mason2012object} includes a large number of rescans but only considers one environment \ie an office.
Moreover, it lacks the appropriate change annotation, so it is unsuitable for calculating metrics. 
Similarly, Ambrus \etal~\cite {ambrus2016unsupervised} capture an office environment with a moving robot. The dataset only records one scene and provides 88 rescans.
We find the annotation of the dataset inconsistent: existing objects in both rescans are falsely annotated as new, while other objects that are physically added in the rescan are not.
The benchmark presented in ~\cite{fehr2017tsdf} uses a hand-held Google Tango device to capture 3 rooms (reference scans) and 23 rescans.
It does not provide any kind of annotation, so the evaluation is much more complicated compared to 3Rscan~\cite{wald2019rio}.
Also, it is much smaller than~\cite{wald2019rio}. Halber \etal~\cite{halber2019rescan} present a method for temporal instance segmentation tracking under changing environments.
Thus, a ground-truth instance annotation is provided for every rescan, but it does not give any information on the changed objects.
The Change-SIM dataset~\cite{park2021changesim} simulates a warehouse environment to evaluate change under non-targeted environmental variations, such as air turbidity.
ChangeSim is a synthetic dataset, whereas 3Rscan is captured under real, challenging conditions.
Finally, Nothing Stands Still (NSS)~\cite{sun2023nothing} is a dataset that focuses on large scenes undergoing large spatial and temporal change.  The dataset is oriented toward 3D registration and does not contain the proper annotation for evaluating change detection.

Considering the lack of available datasets, as detailed above, and following~\cite{adam2022objects}, we evaluate our approach on the 3Rscan dataset. 
The 3Rscan dataset~\cite{wald2019rio} is the richest, real, and most diverse available dataset. The dataset 
is oriented towards instance re-localization and provides information on moved objects and their corresponding rigid transformations.  However, as mentioned in \cref{experimental}, previous work~\cite{adam2022objects} evaluates 3Rscan towards 3D change detection/object discovery by combining the instance segmentation between the reference scan and the rescan and the annotation on moved objects.

\section{Implementation Details}\label{sec:implementation}

\PAR{Segment Anything (SAM).}
SAM~\cite{kirillov2023segment} is the first 2D image segmentation foundation model, working with prompts (\ie, points, bounding boxes, masks, or even text). We exploit this new large, pre-trained network to infer 2D segmentation masks, which help us enforce our “segmentation mask constraint”. Toward the inference of 2D segmentation masks on depth and color images, we deploy the automatic mask generator, as explained in Segment Anything~\cite{adam2022objects}. We use the automatic mask generator by prompting each image with the regular grid of 32×32 points on the full image. As explained in~\cite{kirillov2023segment}, the out-of-the-box automatic mask generator is deployed on another 20 zoomed-in image crops (sourcing from 2x2 and 4x4 partially overlapping windows). The image crops are prompted with points grids of resolution 16x16 and 8x8, respectively.

\PAR{Connected Component Analysis.} When inferring the 2D image masks, objects might be split into multiple segments  (\ie object over-segmentation). As illustrated in \cref{fig:sam}, the couch is split into multiple components in the inferred masks on the color image (\ie the back of the couch, the main part and the arms). Thus, when enforcing our  “segmentation mask constraint” multiple 2D segments for the same object will consequently lead to multiple 3D segments for the same object. To address this challenge, we connect spatially connected segments, \ie we deploy a connected component analysis.  We apply the connected component analysis to the final detected changes. To this end, we compute the connected components on a voxelized representation with the step of 10cm. The step is tuned after examining the resolution of the point cloud so that parts of the same object can indeed be merged into a single object. 

\PAR{Photoconsistency constraint.} The baseline method “photoconsistency constraint”~\cite{taneja2011image} calculates the initial changing regions as our method and as~\cite{palazzolo2018fast, adam2022objects}. It then propagates the changes to all the scene parts with the same texture. Hence, the initial change detection is optimized to enforce color consistency within an object. The optimization is deployed on a voxelized representation of the scene (as described in~\cite{taneja2011image}) via a graph-cut optimization~\cite{boykov2004experimental}.
The binary term of the graph enforces the  “photoconsistency constraint”, and as introduced in~\cite{taneja2011image} is computed as:
\begin{equation} \label{eq:5}
\phi(v_i - v_j) = [v_{i} \neq v_{j}]\cdot\gamma/(\sum{ ||{v_{t}^{i}-v_{t}^{j}}||^2} +1),
\end{equation}
where $||{v_{t}^{i}-v_{t}^{j}}||^2$ accounts for the L2-norm between RGB values of voxels $v_{t}^{i}$ and $v_{t}^{j}$ and $\gamma$ is a regularization factor. 

\begin{figure*}[htb!]
\centering
\includegraphics[width=\linewidth,height=\textheight,keepaspectratio, trim={0cm 7.4cm 10.7cm 0cm},clip]{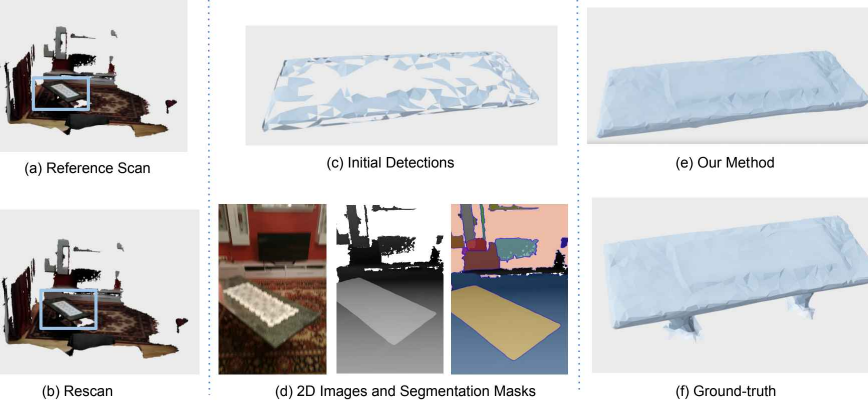} 
\caption{\textbf{A corner case of our method.} The legs of the table are occluded, and no masks are inferred for these regions. Thus, we cannot enforce our “segmentation mask constraint” in these areas and propagate change to them. The discovered object is dense and complete. However, the legs (occluded areas) are missing.}

\label{fig:limitation}
\centering
\end{figure*}

\PAR{Hyperparameters.} The hyperparameters that needed to be tuned for our method are the parameters of the automatic mask generator of SAM~\cite{kirillov2023segment} and the parameters for the graph optimization. Towards the first, we resort to the default parameters as provided by~\cite{kirillov2023segment} since we use SAM out-of-the-box. However, we do show that our method is robust to over-segmentation (\ie the problem that the same object might be split into multiple segments). As shown in \cref{fig:sam}-color, SAM detects multiple segments for each object in the color images (\eg the couch is split into multiple components). On the contrary, SAM leads to fewer and more consistent 2D masks in depth images. Our method performing almost the same for both color and depth images (SAM Color, SAM depth, \cf \cref{Tab:results baselines}) shows its robustness to the over-segmentation challenge. Concerning the hyperparameters for the graph optimization, we tune them as in~\cite{adam2022objects}. 
A more thorough discussion on a sensitivity analysis concerning this set of hyperparameters can be found in \cref{sec:quantitative_suppl}.

\section{Quantitative Results} \label{sec:quantitative_suppl}
\PAR{Sensitivity Analysis.} We perform a sensitivity analysis of our method in relation to the different hyperparameters of the graph optimization. The main hyperparameters that must be tuned are the probabilities for the initial labeling $P$. The initial labeling $P$ will be optimized into $P^{*}$ when enforcing the “segmentation mask constraint”.  The optimized labeling  $P^{*}$  has to remain as close as possible to the initial labeling $P$ (\cf \cref{sec:method}). $P$ encapsulates the initial probability $\rho$ that a supervoxel $v_i$ is labeled as changing $\rho(v_i, l_i = 1)$,  or non-changing  $\rho(v_i, l_i = 0)$. Towards a fair comparison, we set these weights as in~\cite{adam2022objects}:

\begin{align} \label{eq:soft labeling 1}
\rho(v_i, l_i = 1)  = \begin{cases}
0.8 & \text{if changing points $\in  {v_{i}}$} \\
0.5 & \text{if changing points $\notin  {v_{i}}$} \\
\end{cases}\enspace ,\\
\rho(v_i, l_i = 0)  =  \begin{cases}
0.2 & \text{if changing points $\in  {v_{i}}$} \\
0.5 & \text{if changing points $\notin  {v_{i}}$} \\
\end{cases}.
\label{eq:soft labeling 2}\enspace 
\end{align}

However, we also perform a sensitivity analysis to a different set of values, as tabulated in \cref{Tab:results robust}. Through observation of \cref{Tab:results robust}, it is clear that our method is quite robust to that set of hyperparameters, leading to the same results when we assign a high probability to the initially detected changing supervoxels ($\rho(v_i, l_i = 1) \geq 0.7$). When this probability starts to drop, \ie, when we do not trust our initial detections, the performance of our algorithm also seems to drop. The results show the importance of identifying a meaningful set of initial detections, which can be highly trusted during graph optimization. In the case where we do not assign a high probability to our initial detections, the optimized labeling  $P^{*}$ is flexible to diverge from the initial labeling $P$. Thus, “seed supervoxels” (\ie supervoxels containing initial changing points) are removed during the optimization. Consequently, change is not propagated from these regions, and changed objects are not discovered.

\section{Qualitative Results}\label{sec:qualitative_suppl}

\begin{figure*}[htb!]
\centering
\includegraphics[width=\linewidth,height=\textheight,keepaspectratio, trim={0cm 0cm 0cm 0cm},clip]{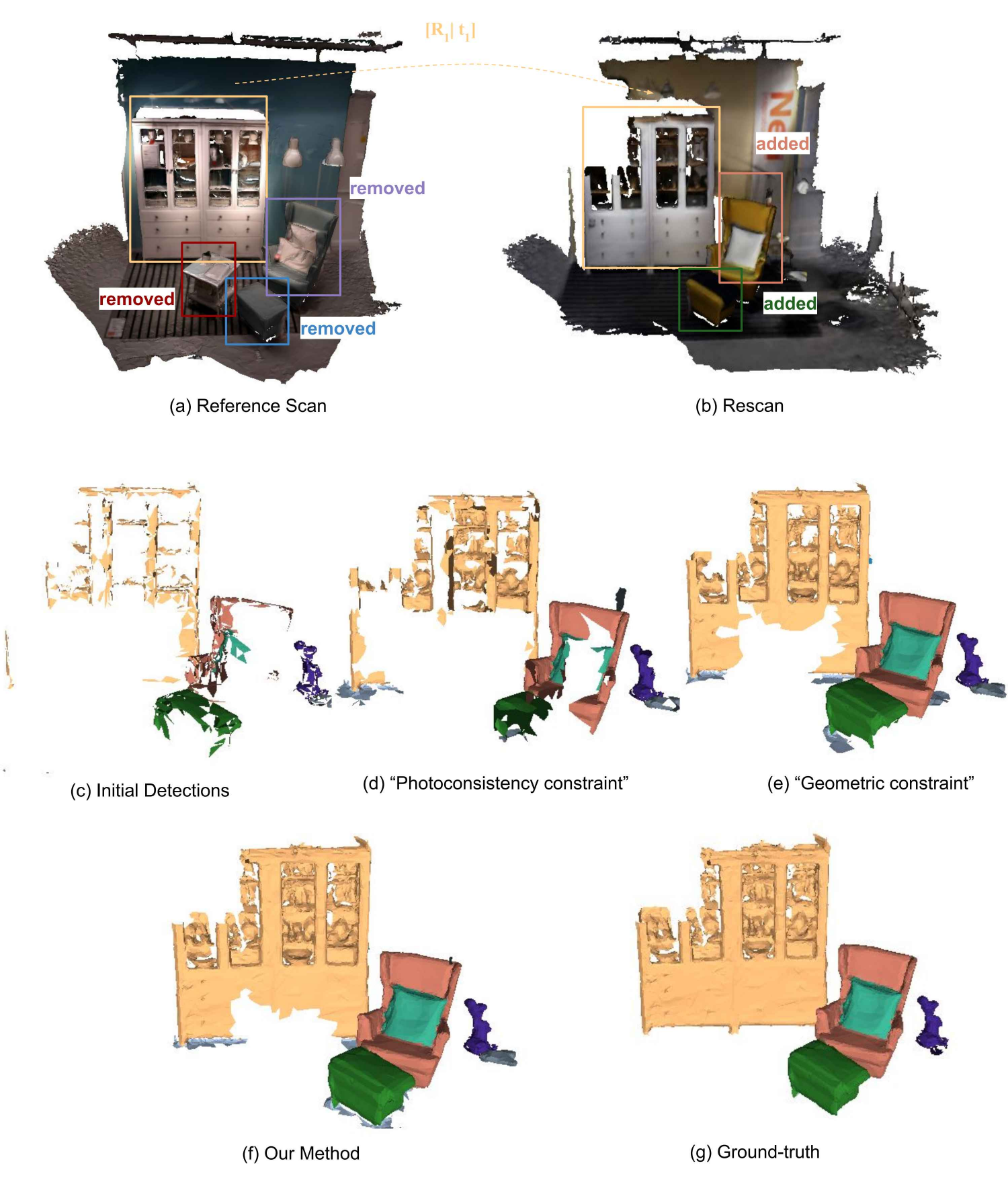} 
\caption{\textbf{Comparative results between the different change propagation constraints.} (a) Reference Scan, (b) Rescan, (c) Initial Detections~\cite{palazzolo2018fast}, (d) “Photoconsistency constraint”~\cite{taneja2011image}, (e) “Geometric constraint”~\cite{adam2022objects}, (f) Our novel “segmentation mask constraint”, and (g) Ground-truth. The objects discovered from the “segmentation mask constraint” are much more complete. The colors used in the visualization come from the instance segmentation of the scene as provided by the dataset.}
\label{fig:all1}
\centering
\end{figure*}

\begin{figure*}[htb!]
\centering
\includegraphics[width=\linewidth,height=\textheight,keepaspectratio, trim={0cm 0cm 0cm 0cm},clip]{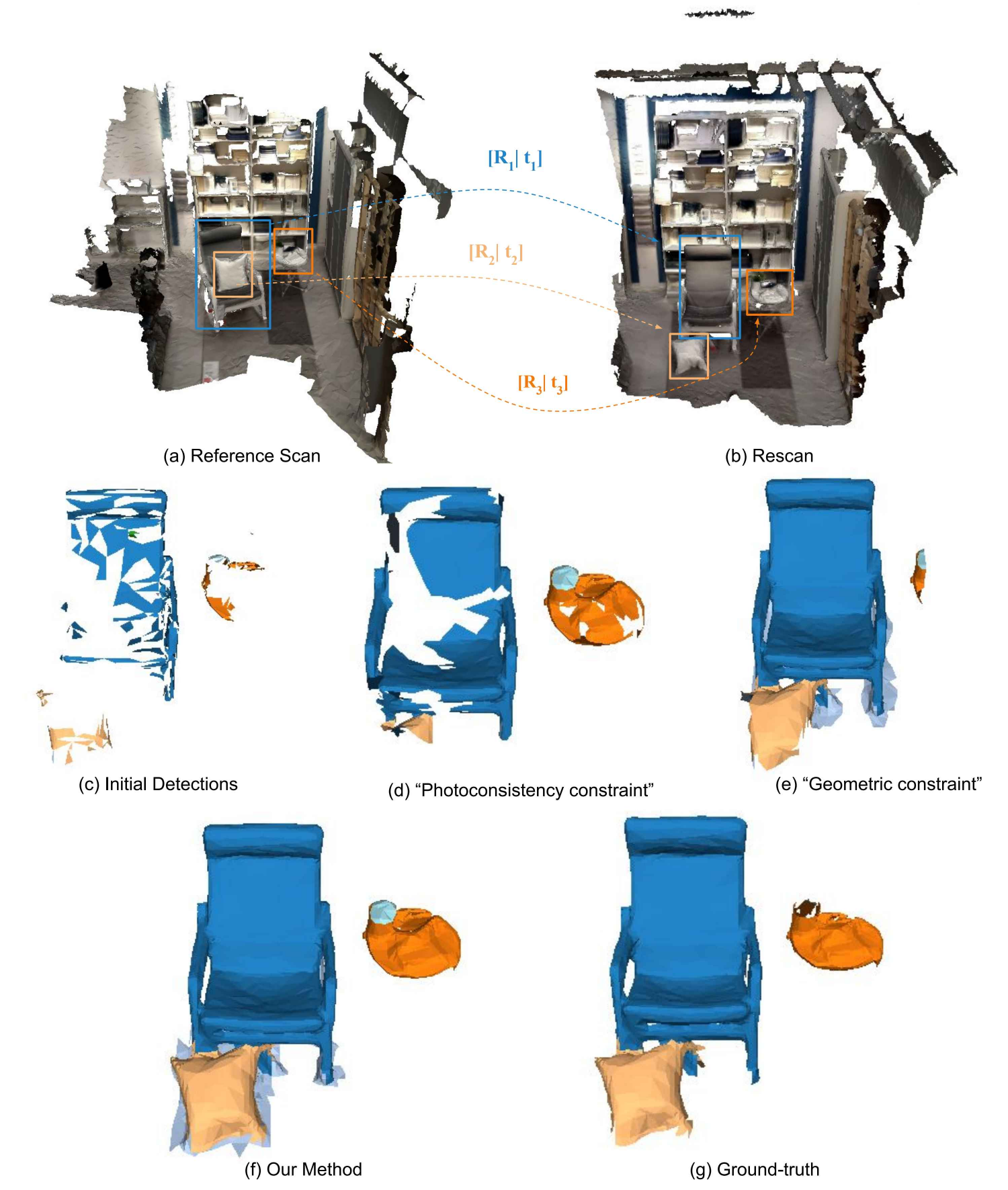} 
\caption{\textbf{Comparative results between the different change propagation constraints.} (a) Reference Scan, (b) Rescan, (c) Initial Detections~\cite{palazzolo2018fast}, (d) “Photoconsistency constraint”~\cite{taneja2011image}, (e) “Geometric constraint”~\cite{adam2022objects}, (f) Our novel “segmentation mask constraint”, and (g) Ground-truth. The objects discovered from the “segmentation mask constraint” are much more complete.  The colors used in the visualization come from the instance segmentation of the scene as provided by the dataset.}
\label{fig:all2}
\centering
\end{figure*}

\cref{fig:all1} and \cref{fig:all2} illustrate more scenes and comparative results. The figures depict the input scans (\ie, reference scan and rescan), results from the initial detection (\ie, equivalent to~\cite{palazzolo2018fast}), results from the “photoconsistency constraint”~\cite{taneja2011image}, results from the “geometric constraint”~\cite{adam2022objects}, results from the proposed method, and the ground-truth changed objects.  It's crucial to emphasize here that in most cases, the rescans represent partial observations of the reference scans. Consequently, the identified changing regions correspond only to the visible areas in the rescan. 

As clear from the \cref{fig:teaser1}, \cref{fig:pipeline}, \cref{fig:all1}, and \cref{fig:all2} the initial detection, which is equivalent to~\cite{palazzolo2017change} leads to rather incomplete results. Palazzoto \& Stachniss~\cite{palazzolo2017change} calculate regions of change by defining inconsistencies in registered images. This published baseline~\cite{palazzolo2017change} can thus be perceived as part of the ablation study since it corresponds to the first step of our method (\ie, initial change detection), ablating the graph optimization. According to the visual results and the metrics in \cref{Tab:results baselines},  it is clear that a more sophisticated, comprehensive solution is needed to tackle this challenging task and that the optimization is critical.

As mentioned in \cref{experimental}, our new “segmentation mask constraint” outperforms a set of competitive baselines. This is also validated by the \cref{fig:all1} and \cref{fig:all2} illustrating that our novel method leads to more dense and complete discovered objects (\eg the cabinet of \cref{fig:all1} and the table of \cref{fig:all2}). The “photoconsistency constraint”~\cite{taneja2011image}  performs well on objects with homogeneous texture, such as the chair depicted in \cref{fig:all2}. However, it is prone to challenging illumination conditions, leading to very sparse discovered objects (\eg the cabinet \cref{fig:all1} ). On the other hand, the “geometric constraint”~\cite{adam2022objects} seems to discover more complete objects than the “photoconsistency constraint”~\cite{taneja2011image}. Nevertheless, it still suffers from establishing noisy rigid transformations that do not propagate the change efficiently to the whole object. This can be attributed to the noisy signal (\eg for the cabinet of \cref{fig:all1} or to deformable objects such as the pillow of \cref{fig:all2}, for which a rigid transformation is difficult to establish). On the other hand, our new “segmentation mask constraint” is conceptually much simpler and does not rely on calculating transformations. Thus, it can handle both rigid and non-rigid objects.

\PAR{Limitations.} In \cref{experimental}, the limitations of our method are discussed. As explained in \cref{experimental}-Limitations, it is clear that our method, as well as~\cite{adam2022objects, taneja2011image, palazzolo2018fast}, suffer from slight misalignments that can lead to wrong initial detections.

Concerning our method, we also present a corner case when a handheld device captures the data. In such cases, all the poses have roughly the same height \ie depending on the human holding the capturing device. Thus, objects or components of objects lying next to the floor (\eg the legs of tables and chairs) are often occluded. In that case, we cannot derive a segmentation mask for those parts of the objects, and thus, we cannot propagate change in these areas. A representative example is depicted in \cref{fig:limitation} where the table has been moved between the two scans. The initial detection is very sparse and only covers parts of the object. Our novel  “segmentation mask constraint”  efficiently propagates the change to the whole table and discovers it effectively. However, since the legs of the table are occluded, they are missing from the final discovered object. 
As mentioned in \cref{experimental}, this could be addressed by generating new poses that are diverse enough to cover the scene from multiple viewpoints. The poses will then be used to render color and depth images, and these images will be used to infer the 2D segmentation masks.


\clearpage

\end{document}